\documentclass[5p,twocolumn]{elsarticle}

\usepackage{amssymb}

\usepackage{hyperref}
\usepackage{url}
\usepackage[cmex10]{amsmath}
\usepackage{amsfonts}
\usepackage{array}
\usepackage{multirow}
\usepackage{caption}
\usepackage{subcaption}
\usepackage[]{algorithm}
\usepackage{verbatim}
\usepackage{perpage}

\usepackage{amsthm}
\usepackage{graphicx}
\usepackage{booktabs}
\usepackage{lineno}
\usepackage[noend]{algpseudocode}
\usepackage{amsmath}

\MakePerPage{footnote} 

\newcommand{\argmax}{\operatornamewithlimits{argmax}}
\newcommand*\rot{\rotatebox{90}}

\journal{arxiv}

\begin{document}

\begin{frontmatter}



\title{Deep Learning for Automatic Stereotypical Motor Movement Detection using Wearable Sensors in Autism Spectrum Disorders}

\author{Nastaran Mohammadian Rad \corref{cor1} \fnref{aff1,aff2,aff3}} \ead{nastaran@fbk.eu}
\author{Seyed Mostafa Kia \fnref{aff4,aff5}}
\author{Calogero Zarbo \fnref{aff1}}
\author{Twan van Laarhoven \fnref{aff2}} 
\author{Giuseppe Jurman \fnref{aff1}} 
\author{Paola Venuti \fnref{aff6}}
\author{Elena Marchiori \fnref{aff2}}
\author{Cesare Furlanello \fnref{aff1}}

\fntext[aff1]{Fondazione Bruno Kessler, Trento, Italy}
\fntext[aff2]{Institute for Computing and Information Sciences, Radboud University, Nijmegen, The Netherlands}
\fntext[aff3]{Department of Information Engineering and Computer Science, University of Trento, Trento, Italy}
\fntext[aff4]{Donders Centre for Cognitive Neuroimaging, Donders Institute for Brain, Cognition and Behaviour, Radboud University, Nijmegen, The Netherlands}
\fntext[aff5]{Department of Cognitive Neuroscience, Radboud University Medical Centre, Nijmegen, The Netherlands}
\fntext[aff6]{Department of Psychology and Cognitive Science, University of Trento, Trento, Italy}

\cortext[cor1]{Corresponding author}
\address{Fondazione Bruno Kessler, Via Sommarive 18, 38123, Povo, Trento, Italy}

\begin{abstract}
Autism Spectrum Disorders are associated with atypical movements, of which stereotypical motor movements (SMMs) interfere with learning and social interaction. The automatic SMM detection using inertial measurement units (IMU) remains complex due to the strong intra and inter-subject variability, especially when handcrafted features are extracted from the signal. We propose a new application of the deep learning to facilitate automatic SMM detection using multi-axis IMUs. We use a convolutional neural network (CNN) to learn a discriminative feature space from raw data. We show how the CNN can be used for parameter transfer learning to enhance the detection rate on longitudinal data. We also combine the long short-term memory (LSTM) with CNN to model the temporal patterns in a sequence of multi-axis signals. Further, we employ ensemble learning to combine multiple LSTM learners into a more robust SMM detector. Our results show that: 1) feature learning outperforms handcrafted features; 2) parameter transfer learning is beneficial in longitudinal settings; 3) using LSTM to learn the temporal dynamic of signals enhances the detection rate especially for skewed training data; 4) an ensemble of LSTMs provides more accurate and stable detectors. These findings provide a significant step toward accurate SMM detection in real-time scenarios.
\end{abstract}

\begin{keyword}
Convolutional Neural Networks \sep Long Short-Term Memory \sep Transfer Learning \sep Ensemble Learning \sep Wearable Sensors \sep Autism Spectrum Disorders
\end{keyword}

\end{frontmatter}


\section{Introduction}
\label{sec:introduction}
Autism spectrum Disorder (ASD) is a complex and heterogeneous neuro-developmental disorder. Specific symptoms in ASD are difficulties in social interactions, repetitive or restricted behaviors, and verbal/nonverbal communication difficulties. Prevalence of ASD is reported to be 1 in 88 individuals~\cite{baio2012prevalence}. While the majority of studies have mainly focused on social and communication problems of ASD children, the repetitive and restricted behaviors associated with ASD individuals are also objects of interest because of their effect on the learning performance and socialization; and also as an indicator of distress~\cite{hernandez2014using, chaspari2012interplay, hedman2012measuring}. Stereotypical Motor Movements (SMMs) are the major group of atypical repetitive behaviors in children with ASD. SMMs occur without evoking stimuli and include hand flapping, body rocking, and mouthing~\cite{berkson1962stereotyped, lewis1982stereotyped, ridley1982stereotypy}. SMMs have a specific negative effect on the quality of life of ASD children: they can affect negatively on the performance of children during learning new skills and while using the learned skills~\cite{varni1979analysis}. Furthermore, since these type of movements are socially abnormal, they cause difficulties in interaction with pairs in the school or other social settings~\cite{jones1990social}. In some cases, the severity of SMMs can lead to a meltdown event and even can cause self-damaging behaviors~\cite{kennedy2002evolution}.

There are three traditional approaches for measuring the SMMs~\cite{sprague1996stereotyped}: 1) paper-and-pencil rating, 2) direct behavioral observation, and 3) video-based methods. Paper-and-pencil rating is an interview-based approach which suffers from the subjectivity in rating. Furthermore, it cannot accurately detect the intensity, amount, and duration of SMMs~\cite{pyles1997stereotypy}. In the direct behavioral observation approach, therapists can directly observe and record the sequences of SMMs. This method is  not also a reliable approach due to the several reasons~\cite{sprague1996stereotyped, gardenier2004comparison}: first, in high speed movements, it is hard for therapists to accurately observe and document all SMM sequences instantaneously. Second, determining the start and end time of the SMM sequences is difficult. Third, it is impossible for therapists to concurrently record all environmental conditions and SMMs. Video-based approaches are based on video capturing, offline coding, and analysis of SMMs. Since multiple coding sessions of the captured videos are feasible, this observational method is much more accurate than two previous approaches, but it is time-consuming and unpractical as a clinical tool\cite{matson2007assessing}. 

\begin{figure*}[t]
  \centering
  \includegraphics[width=0.95\textwidth]{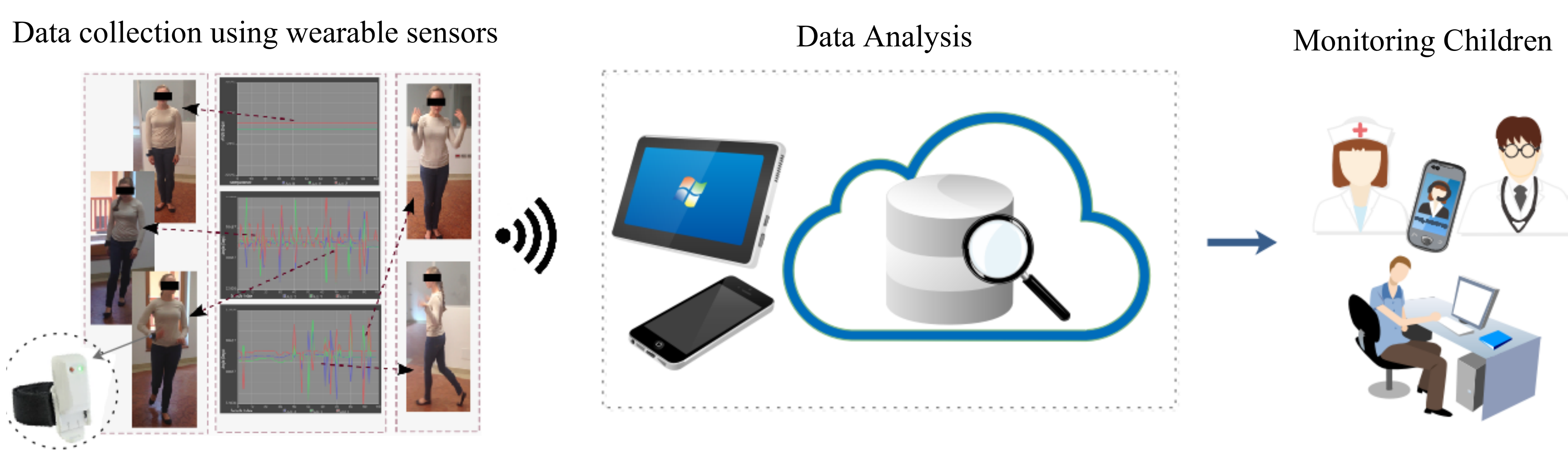}
  \caption{A real-time automatic SMM detection system. Inertial Measurement Units (IMUs) can be used for data collection. The collected data can be analyzed locally or remotely to detect SMMs. In case of detecting abnormal movements, an alert is sent to a therapist, caregiver, or parents.}
  \label{fig:SMM_Detection} 
\end{figure*}

Considering the high prevalence rate of autism in children~\cite{goldman2009motor} and the limitations of existing methods for measuring SMMs, it is essential to develop time efficient and accurate methods for automatic SMM detection. Developing a real-time SMM detection and quantification system would be advantageous for ASD researchers, caregivers, families, and therapists. Such a system would provide a powerful tool to evaluate the adaptation of subjects with ASD to diverse life contexts within an ecologic approach. In particular, it helps to mitigate the meltdown behaviors that are anticipated by the increase in atypical behaviors. Any automatic quantification of atypical movements would indeed help caregivers and teachers to defuse the mechanism leading to stereotyped behaviors by involving children in specific activities or social interactions. Such involvement decreases the frequency of SMMs and gradually alleviates their duration and severity~\cite{lee2007social, loftin2008social}. A real-time implementation of SMM detection system (see Figure~\ref{fig:SMM_Detection}) would help therapists to evaluate the efficacy of behavioral interventions.

Inertial Measurement Units (IMUs) provide effective tools for measuring the frequency, intensity, and duration of physical activities over a time period via embedded accelerometer, gyroscope, and magnetometer sensors. Due to the small size and possibility of embedding in the mobile phones, IMUs have been adopted as common and useful sensors for wearable devices to measure the physical activities in either constrained and free-living environments~\cite{mathie2004accelerometry,lara2013survey,helbostad2017mobile}. In recent years, human activity recognition using IMU sensors has been widely studied. Most of these studies tried to extract time and frequency domain features such as mean, standard deviation, skewness, and FFT peaks from raw signals to feed them to a classifier for activity identification~\cite{baek2004accelerometer,gjoreski2016how}. According to the achieved results in human activity recognition systems, applying pattern recognition on the collected data by IMU sensors can reliably and accurately detect physical activities which are an evidence for the possibility of applying such techniques to automatically detect SMMs in ASD children~\cite{westeyn2005recognizing,min2009optimal,min2010automatic,min2010novel,gonccalves2012automatic,gonccalves2012automatic1,
rodrigues2013stereotyped,albinali2012detecting,albinali2009recognizing,goodwin2011automated,goodwin2014moving,
plotz2012automatic,Großekathöfer2017Automated,Amiri2017WearSense}. Despite meaningful amount of research in this direction, few challenges for automatic SMM detection using wearable sensors still remain unsolved especially in real-time applications.

One of the important challenges for accurate SMM detection is to extract a set of effective and robust features from the IMU signal. As in many other signal processing applications, SMM detection is commonly based on extracting handcrafted features from the IMU signals. So far, a wide variety of feature extraction methods have been used in the literature. Generally two main types of features are extracted from the accelerometer signal~\cite{ince2008detection}: i) time domain features, ii) frequency domain features. For time domain features, some statistical features such as mean, standard deviation, zero-crossing, energy, and correlation are extracted from the overlapping windows of the signal. In the case of frequency features the discrete Fourier transform is used to estimate the power of different frequency bands. In addition, the Stockwell transform~\cite{stockwell1996localization} is proposed by~\cite{goodwin2014moving} for feature extraction from 3-axis accelerometer signals in order to provide better time-frequency resolution for non-stationary signals. In spite of their popularity, manual feature extraction and selection suffer from subjectivity and time inefficiency~\cite{martinez2013learning} that restrict the performance and also the application of SMM detection systems in real-time scenarios. 

Another challenge toward developing a real-time SMM detection system is personalization due to the intra and inter-subject variability~\cite{goodwin2014moving,Großekathöfer2017Automated}. This challenge, despite its crucial importance, has been undervalued~\cite{goodwin2014moving}. Intra-subject variability is mainly due to the high variability in the intensity, duration, frequency, and topography of SMMs in each individual with ASD. Inter-subject differences are defined by the same variability across different individuals. Existence of these two types of variability within and across ASD persons motivates the necessity of developing an adaptive SMM detection algorithm that is capable to adjust to new patterns of behaviors. Feature learning and transfer learning~\cite{pan2010survey} can be considered as candidate solutions to attack these challenges. To this end, here we present an extended version of our previous efforts in~\cite{rad2016applying, rad2016stereotypical} with four main contributions: 1) robust feature learning from multi-sensor IMU signals; 2) enhancing the adaptability of SMM detection system to new data via parameter transfer learning; 3) improving the detection rate by incorporating the temporal dynamics of signals in the feature learning process; and 4) using principles of the ensemble learning to enhance the detection rate. 

To achieve our first goal, we propose a new application of the deep learning paradigm in order to directly \emph{learn} discriminating features for detecting SMM patterns. In particular, we use a convolutional neural network (CNN)~\cite{lecun1995convolutional} to bypass the commonly used feature extraction procedure. The idea of the CNN is inspired by the visual sensory system of living creatures~\cite{fukushima1980neocognitron}. Following this idea, LeCun et al.~\cite{lecun1998gradient} developed a deep CNN architecture to address a pattern recognition problem in computer vision. Having fewer connections and parameters due to the weight sharing property, CNNs are easier to train compared to other deep neural networks. Currently, CNN solutions are among the best-performing systems on pattern recognition systems specifically for the handwritten character~\cite{lecun1998gradient} and object recognition~\cite{krizhevsky2012imagenet}. Beyond audio and image recognition systems, CNNs are successfully applied on various types of signals. Mirowski et al.~\cite{mirowski2008comparing} applied CNN on EEG signals for seizure detection. In the domain of psychophysiology, for the first time Martinez et al.~\cite{martinez2013learning} proposed a model based on CNN to predict affective states of fun, excitement, anxiety, and relaxation. Their proposed model was tested on skin conductance and blood volume pulse signals. Recent studies show the advantageous of applying CNN on accelerometer signals for human activity recognition~\cite{yang2015deep, zeng2014convolutional,Zebin2016Human}. 

To fulfill our second goal, we employ the parameter transfer learning by pre-initializing the parameters of the CNN~\cite{morales2016deep}. We hypothesize that this capability can be used to transfer the prior knowledge regarding the distribution of parameters from one dataset to another dataset that are collected in a longitudinal study. If successful, our method can be employed in order to enhance the adaptability of SMM detection system to new unseen data, thus facilitates its applications in wild real-world scenarios. 

By the definition, SMMs are repetitive behaviors, thus temporal patterns stored in the sequence of samples are expected to contain valuable information. Our third contribution relies on the fact that the proposed CNN architecture does not fully exploit the temporally structured information in the IMU signal sequences. This is a general issue in SMM detection, in which the segments of IMU signals are treated as statistically independent samples. Therefore the possible long-term dependencies stored in the longer temporal intervals of the signal are ignored in the detection process. Long short-term memory (LSTM)~\cite{hochreiter1997long} as a type of recurrent neural networks (RNN) has been effectively used for learning long-term temporal dependencies in sequential data such as speech recognition~\cite{eyben2009speech}, handwriting recognition~\cite{doetsch2014fast}, and natural language modeling~\cite{zaremba2014recurrent}. Recently, the LSTM has also been successfully used for human activity recognition using wearable technologies as a classic sequence analysis problem~\cite{ordonez2016deep,hammerla2016deep,Zebin2016Human}. Considering these studies in human activity recognition, it is expected that learning the temporal patterns stored in the consecutive samples of IMU data to provide higher accurate SMM detectors. Thus, here we propose a deep architecture, stacking an LSTM layer on top of the CNN architecture, in order to learn the dynamic feature space on the sequence of IMU data. We further show that combining multiple LSTM models using an ensemble learning technique can improve the stability of results. To the best of our knowledge, it is the first time that a recurrent architecture is used for SMM detection using wearable technologies.

The rest of this paper is organized as follows: in Section~\ref{sec:methods} we first briefly review the formal definitions and concepts about CNN, LSTM, parameter transfer learning, and the ensemble of the best base learners approach. Then using the presented definitions we introduce the proposed CNN and LSTM architectures~\footnote{Here the architecture implies the customization of the structural parameters of the CNN such as the number of layers, the number and size of filters, etc.} for SMM detection on IMU signals. The experimental materials and procedures are explained in this section. Section~\ref{sec:results} compares our experimental results versus the state of the art solutions in SMM detection. Our results on a simulated dataset and two real datasets show that feature learning via the CNN outperforms handcrafted features in SMM classification. Furthermore, it is shown that the parameter transfer learning is beneficial in enhancing the SMM detection rate when moving to a new dataset. Finally our results illustrate that including the dynamics of recorded data in feature learning process improves the classification performance in SMM detection especially when an unbalanced training set is used in the training phase. In Section~\ref{sec:discussion} we discuss how the proposed deep architecture facilitates developing real-time SMM detection systems. We further discuss the main limitation of our method and state the possible future directions. Section~\ref{sec:conclusions} concludes this paper by summarizing our achievements.
\section{Methods}
\label{sec:methods}
\subsection{Notation}
\label{subsec:notation}
Let $\textbf{S}_1,\textbf{S}_2,\dots,\textbf{S}_c \in \mathbb{R}^{L}$ be $c$ time-series of length $L$ that are recorded by a set of inertial measurement units (IMUs) (e.g., accelerometer, gyroscope, and magnetometer sensors) at the sampling frequency of $\nu$ Hz. Thus, $T=L/ \nu$ represents the length of the signal in seconds. We refer to each $\textbf{S}_i$ as a data channel. Now consider $\textbf{X}_t \in \mathbb{R}^{c \times \nu}$ for $t \in \{ 1,2,\dots,T\}$ as a sample in the raw feature space that is constructed by concatenating time-series of $c$ data channels in a given time $t$. Let $y_t \in \{0,1\}$ be the label associated to $\textbf{X}_t$ where $y_t = 1$ corresponds to an SMM sample. In this text, we use boldface capital letters to represent matrices, boldface lowercase letters to represent vectors, and italic lowercase letters to represent scalars. We represent the matrix and element-wise multiplication between $\textbf{A}$ and $\textbf{B}$ matrices by $\textbf{A} \cdot \textbf{B}$ and $\textbf{A} \odot \textbf{B}$, respectively. Further, $[\textbf{a},\textbf{b}]$ represents vector concatenation operation between $\textbf{a}$ and $\textbf{b}$ vectors.

\subsection{Feature Learning via Convolutional Neural Network}
\label{subsec:CNN}
The goal of an SMM detector is to predict the probability of being an SMM for a given sample $\textbf{X}_t$, i.e., $P(y_t=1 \mid \textbf{X}_t)$. While the raw feature space ($\textbf{X}_t$) is sensitive to intra and inter-subject differences, feature learning can provide the possibility to learn a new feature space that is robust over time and across different subjects. The aim of \emph{feature learning} is to learn a linear or non-linear mapping function $\mathcal{F}:\textbf{X}_t \mapsto \textbf{x}'_t$, where $\textbf{x}'_t \in \mathbb{R}^d$ is called the learned feature space. Then a classifier can be used in the learned feature space to estimate $P(y_t=1 \mid \textbf{x}'_t)$. 

Convolutional neural networks (CNNs) offer an effective infrastructure for feature learning. CNN benefits from invariant local receptive fields, shared weights, and spatio-temporal sub-sampling features to provide robustness over shifts and distortions in the input space~\cite{lecun1995convolutional}. A classic CNN has a hierarchical architecture that alternates convolutional and pooling layers in order to summarize large input spaces with spatio-temporal relations into a lower dimensional feature space. A 1D-convolutional layer receives the input signal $\textbf{X}_t \in \mathbb{R}^{c \times \nu}$, convolves it with a set of $f$ filters with the length of $m$, $\textbf{W} \in \mathbb{R}^{f \times c \times m}$, and produces a feature map $\textbf{M}_t \in \mathbb{R}^{f \times \nu}$:
\begin{eqnarray} \label{eq:convolutional_layer}
\begin{split}
& \textbf{M}_t = \textbf{X}_t * \textbf{W} = \\ 
& \resizebox{0.4\textwidth}{!}{$\begin{bmatrix}
\sum_{j=1}^c \sum_{i=1}^m w_{1,j,i} \times x_{j,1+i} & \dots & \sum_{j=1}^c \sum_{i=1}^m w_{1,j,i} \times x_{j,\nu+i}\\ 
\sum_{j=1}^c \sum_{i=1}^m w_{2,j,i} \times x_{j,1+i} &  \dots & \sum_{j=1}^c \sum_{i=1}^m w_{2,j,i} \times x_{j,\nu+i} \\ 
\vdots  & \vdots & \vdots \\ 
\sum_{j=1}^c \sum_{i=1}^m w_{f,j,i} \times x_{j,1+i} &  \dots & \sum_{j=1}^c \sum_{i=1}^m w_{f,j,i} \times x_{j,\nu+i}
\end{bmatrix}$}
\end{split}
\end{eqnarray}
where $*$ represents the convolution operator. The feature map is then fed to an activation function, generally the rectified linear unit (ReLU), to add non-linearity to the network and also to avoid the gradient vanishing problem~\cite{Goodfellow2016Deep}, where:
\begin{eqnarray} \label{eq:relu_function}
\textbf{M}_t^+ = max(\textbf{0}^{f \times \nu},\textbf{M}_t).
\end{eqnarray}
Here $max(.,.)$ represents the element-wise max operation. Finally, in order to reduce the sensitivity of the output to shifts and distortions, $\textbf{M}_t^+$ is passed through a pooling layer which performs a local averaging or sub-sampling over a pooling window with size of $p$ elements and calculates the reduced feature map $\textbf{M}'_t \in \mathbb{R}^{f \times \frac{\nu}{u}}$. In fact, a pooling layer reduces the resolution of a feature map by factor of $\frac{1}{u}$ where $u$ is the stride (or step) size. Max-pooling and average-pooling are two commonly used pooling functions which compute the maximum or average value among the values in a pooling window, respectively. In average-pooling for $m'_{i,j} \in \textbf{M}'_t$, $i \in \{1, \dots, f\}$, and $j \in \{1, \dots, \frac{\nu}{u}\}$, we have:
\begin{eqnarray} \label{eq:avg_pooling}
m'_{i,j} = \frac{1}{p} \sum_{k=1}^p m_{i,(j-1)\times u + k}.
\end{eqnarray}

Alternatively, in the max-pooling each element of the reduced feature map is the maximum value in a corresponding pooling window:
\begin{eqnarray} \label{eq:max_pooling}
\resizebox{0.4\textwidth}{!}{$m'_{i,j} =  max(m_{i,(j-1)\times u + 1},m_{i,(j-1)\times u + 2},\dots, m_{i,(j-1)\times u + p}).$}
\end{eqnarray}

The reduced feature map $\textbf{M}'_t$ can be used as the input to the next convolutional layer, i.e., $\textbf{X}_t$ of the next layer. In general, the reduced feature map computed by stacking several convolution, ReLU, and pooling layers is flattened as a vector before the classification step. The flattening step is performed by collapsing the rows of $\textbf{M}'_t$ in the form of a vector. The resulting vector is called the learned feature space $\textbf{x}'_t$ that represents a new representation of the original feature space. This new representation is typically fed to a fully connected neural network followed by a softmax layer for the classification purposes. 

\begin{figure*}[t]
  \centering
  \includegraphics[width=0.95\textwidth]{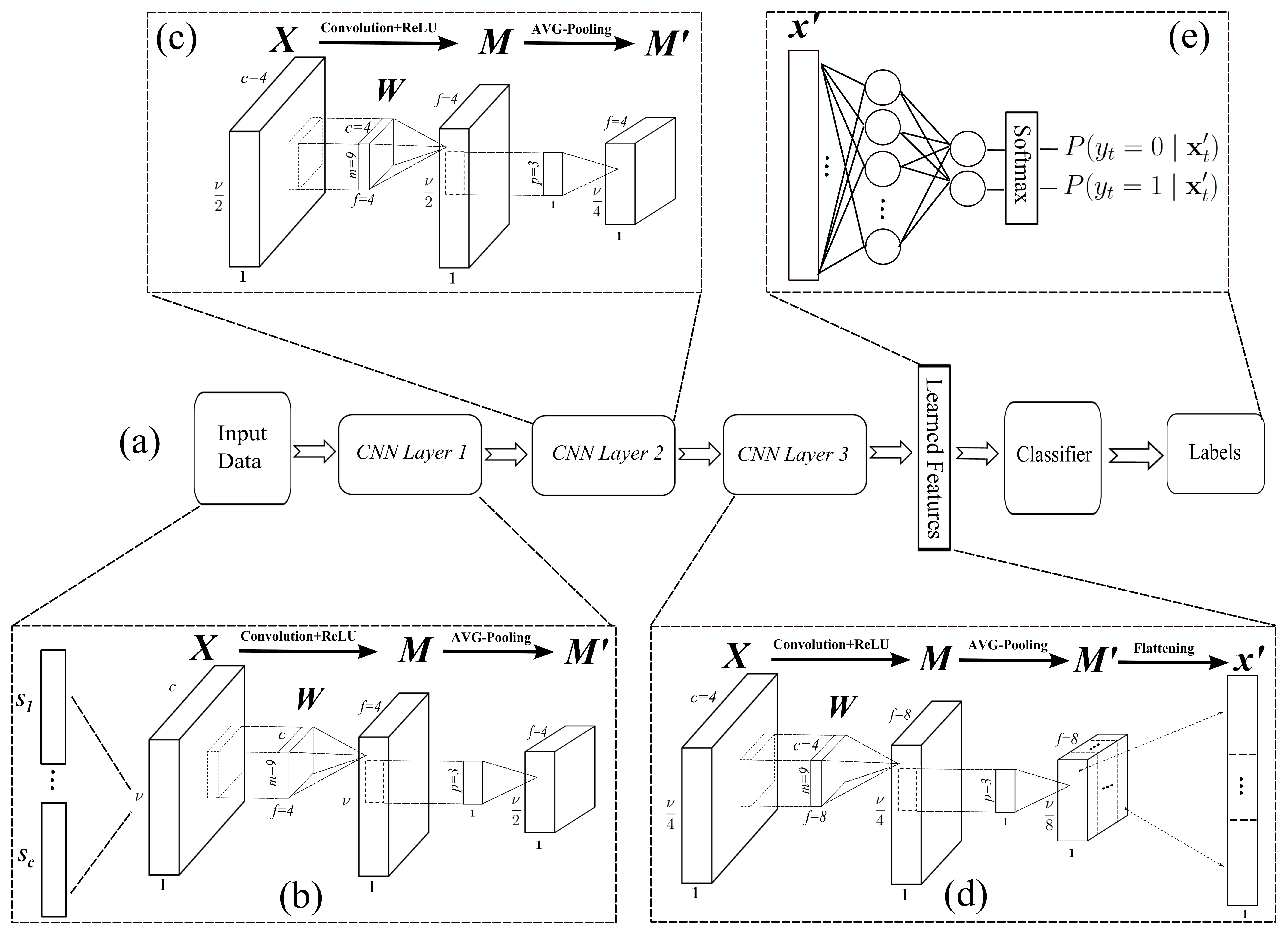}
  \caption{(a) The proposed architecture for SMM detection in the static feature space using a three-layer CNN. (b) The first CNN layer. This layer receives the one-second long time-series of several IMU sensors at time $t$, i.e., $\textbf{X}_t$, and transfer it to the first level reduced feature map $\textbf{M}'_t$. (c) The second CNN layer that uses the first level reduced feature map $\textbf{M}'_t$ as its input, and transfer it to the second-level reduced feature map. (d) The third CNN layer. The reduced feature map of this layer is reshaped to the learned feature vector $\textbf{x}'_t$ using the flattening operation. (e) The learned feature vector is fed to a fully-connected followed by a softmax layer to classify the samples to SMM and no-SMM classes.}
  \label{fig:CNN_Architecture} 
\end{figure*}

In this paper, and for the purpose of SMM detection on the multi-sensor IMU data, we propose to use a three-layer CNN to transform the time-series of multiple sensors to a new feature space. The proposed architecture is shown in Figure~\ref{fig:CNN_Architecture}. Three convolutional layers are designed to have $4$, $4$, and $8$ filters with the length of $9$ samples (i.e., $0.1$ seconds), respectively. The length of the pooling window and pooling stride are fixed to $3$ ($p=3$) and $2$ ($u=2$), respectively. The pooling stride of $2$ reduces the length of feature maps by the factor of $0.5$ after each pooling layer. The output of the third convolutional layer after flattening provides the learned feature vector. Then, the learned feature vector is fed to a two-layer fully-connected network with $8$ and $2$ neurons that are connected to a softmax layer. A dropout~\cite{srivastava2014dropout} rate of $0.5$ is used in the fully connected layers to avoid the overfitting problem. Since only the information in $\textbf{X}_t$ is used to compute $\textbf{x}'_t$ and then predict $y_t$, we refer to the learned feature space via this CNN architecture as the static feature space.

\subsection{Parameter Transfer Learning via Network Pre-initialization}
\label{subsec:transfer_learning}
The quality and characteristics of recorded IMU signals varies not only from subject to subject but also from time to time in a single subject. Therefore it is important that the SMM detector system be able to adapt to new streams of signals in longitudinal scenarios. In this paper, we explore the possibility of parameter transfer learning via network pre-initialization in order to transfer the learned patterns to the newly seen data in a different time span. In this direction we first formalize the background theoretical concepts.

In the statistical learning theory, the goal is to learn a task $\mathcal{T}$ in a certain domain $\mathcal{D}$. A \emph{domain} $\mathcal{D}=\{\mathcal{X},\rho_{\mathcal{X}}\}$ is defined as a possible conjunction between an input space $\mathcal{X}$ and a marginal probability distribution $\rho_{\mathcal{X}}$. For example in the SMM detection context, the recorded IMU signal for different subjects can be considered as different domains as the marginal probability distribution $\rho_{\mathcal{X}}$ is different from one subject to another. Similarly different domains can be defined by time in longitudinal data collection scenarios. Given a domain $\mathcal{D}$, a \emph{task} $\mathcal{T}=\{\mathcal{Y},\Phi\}$ is defined as a predictive function $\Phi$ from $\mathcal{D}$ to the output space $\mathcal{Y}$. For example in this study $\Phi$ is the SMM detector, and $\mathcal{Y}$ represents the categorical output space of SMM and no-SMM samples. Assume $\mathcal{D}_S$, $\mathcal{D}_T$, $\mathcal{T}_S$, and $\mathcal{T}_T$ to represent the source domain, target domain, source task, and target task, respectively. \emph{Transfer learning} aims to benefit from the knowledge in the source domain and task in order to improve the predictive power in the target domain when $\mathcal{D}_S \neq \mathcal{D}_T$ or $\mathcal{T}_S \neq \mathcal{T}_T$~\cite{pan2010survey}. 

In this study, we are interested in the application of \emph{parameter transfer learning} via pre-initializing the parameters of a deep neural network, as a well-established technique in the deep learning community, to improve the SMM prediction performance across different subjects and time intervals. To this end, we define the source domain $\mathcal{D}_S$ as the IMU signal which is collected on several subjects at the time span $T_1$. Similarly the target domain $\mathcal{D}_T$ is defined as the IMU signal which is collected on several subjects at the time span $T_2$. Assume $\Phi_S$ be the learned predictive function, i.e., the CNN classifier, in the source domain. We use the learned parameters in $\Phi_S$ to pre-initialize the parameters of the predictive function in the target domain $\Phi_T$. In simpler words, instead of random pre-initialization, we initialize the parameters of CNN classifier in the target domain with the learned CNN parameters in the source domain. We hypothesize that such a knowledge transfer via learned parameters improves the prediction performance in the longitudinal studies where the data are collected at different time intervals.

\subsection{SMM Detection in Dynamic Feature Space using LSTM}
\label{subsec:dynamic_SMM_detection}
In SMM detection using static feature space (see Section~\ref{subsec:CNN}) only the data in $\textbf{X}_t$ is used to predict $y_t$. Thus it is implicitly assumed that the sequence of samples over time are independent and identically distributed (i.i.d). But in reality, this assumption is not valid as the samples in consecutive time steps are highly dependent. Therefore, it is expected that accounting for this dependency would improve the performance of the SMM detector. Following this hypothesis, we propose to use a long short-term memory (LSTM) layer to model the temporal dependency between the consecutive time steps of the recorded signal.

Let $\textbf{x}'_t$ be a set of static features that are extracted or learned from samples in the raw feature space (i.e., from $\textbf{X}_t$). Here we assume the CNN architecture explained in Section~\ref{subsec:CNN} is used to compute $\textbf{x}'_t$. Then, let $\textbf{c}_{t} \in \mathbb{R}^q$ and $\textbf{h}_t \in \mathbb{R}^q$ to represent the cell state and output of an LSTM unit at time step $t$, respectively, where $q$ is the number of neurons in the LSTM unit. We will refer to $\textbf{h}_t$ as the \emph{dynamic} feature space. The LSTM unit receives $\textbf{x}'_t$, $\textbf{h}_{t-1}$, and $\textbf{c}_{t-1}$ as its inputs, and computes $\textbf{c}_t$ and $\textbf{h}_t$ as follows:
\begin{eqnarray} \label{eq:state_gate}
\textbf{c}_t = \textbf{f}_t \odot \textbf{c}_{t-1} + \textbf{i}_t \odot \tilde{\textbf{c}}_t ,
\end{eqnarray}
\begin{eqnarray} \label{eq:output}
\textbf{h}_t = \textbf{o}_t \odot (1 - e ^{-2 \times \textbf{c}_t}) \odot (1 + e ^{-2 \times \textbf{c}_t})^{-1} .
\end{eqnarray}

Here $\textbf{f}_t \in \mathbb{R}^q$ is called the forget gate vector and its elements are real numbers between $0$ and $1$ that decide how much information to be passed from $\textbf{c}_{t-1}$ to $\textbf{c}_{t}$. During the learning phase, the forget gate learns the forget weight matrix $\textbf{W}_\textbf{f}$ and the forget bias vector $\textbf{b}_\textbf{f}$. $\textbf{f}_t$ is computed by
\begin{eqnarray} \label{eq:forget_gate}
\textbf{f}_t = (1+e^{-(\textbf{W}_\textbf{f} \cdot [\textbf{h}_{t-1},\textbf{x}'_t]+\textbf{b}_\textbf{f})})^{-1} .
\end{eqnarray}

Using a tangent hyperbolic function, $\tilde{\textbf{c}}_t \in \mathbb{R}^q$ provides new candidate values between $-1$ and $1$ for $\textbf{c}_t$ by learning $\textbf{W}_\textbf{c}$ and $\textbf{b}_\textbf{c}$: 
\begin{eqnarray} \label{eq:semi_state_gate}
\begin{split}
\resizebox{0.4\textwidth}{!}{$\tilde{\textbf{c}}_t = (1-e^{-2 \times(\textbf{W}_\textbf{c} \cdot [y_{t-1},\textbf{x}'_t]+\textbf{b}_\textbf{c})})\odot (1+e^{-2 \times(\textbf{W}_\textbf{c} \cdot [y_{t-1},\textbf{x}'_t]+\textbf{b}_\textbf{c})})^{-1} ,$}
\end{split}
\end{eqnarray}
where $\textbf{i}_t \in \mathbb{R}^q$ is the input gate vector with elements between $0$ and $1$. These values determine the level of new information in $\tilde{\textbf{c}}_t$ to be transferred to the cell state $\textbf{c}_{t}$. $\textbf{i}_t$ is computed based on $\textbf{W}_\textbf{i}$ and $\textbf{b}_\textbf{i}$ as follows:
\begin{eqnarray} \label{eq:input_gate}
\textbf{i}_t = (1+e^{-(\textbf{W}_\textbf{i} \cdot [y_{t-1},\textbf{x}'_t]+\textbf{b}_\textbf{i})})^{-1} .
\end{eqnarray}

Finally, $\textbf{o}_t \in \mathbb{R}^q$ is the output gate vector that filters the cell state $\textbf{c}_t$ to generate the output of the LSTM unit $\textbf{h}_t$:
\begin{eqnarray} \label{eq:output_gate}
\textbf{o}_t = (1+e^{-(\textbf{W}_\textbf{o} \cdot [y_{t-1},\textbf{x}'_t]+\textbf{b}_\textbf{o})})^{-1} .
\end{eqnarray}

\begin{figure}[t]
         \centering
         \includegraphics[width=0.45\textwidth]{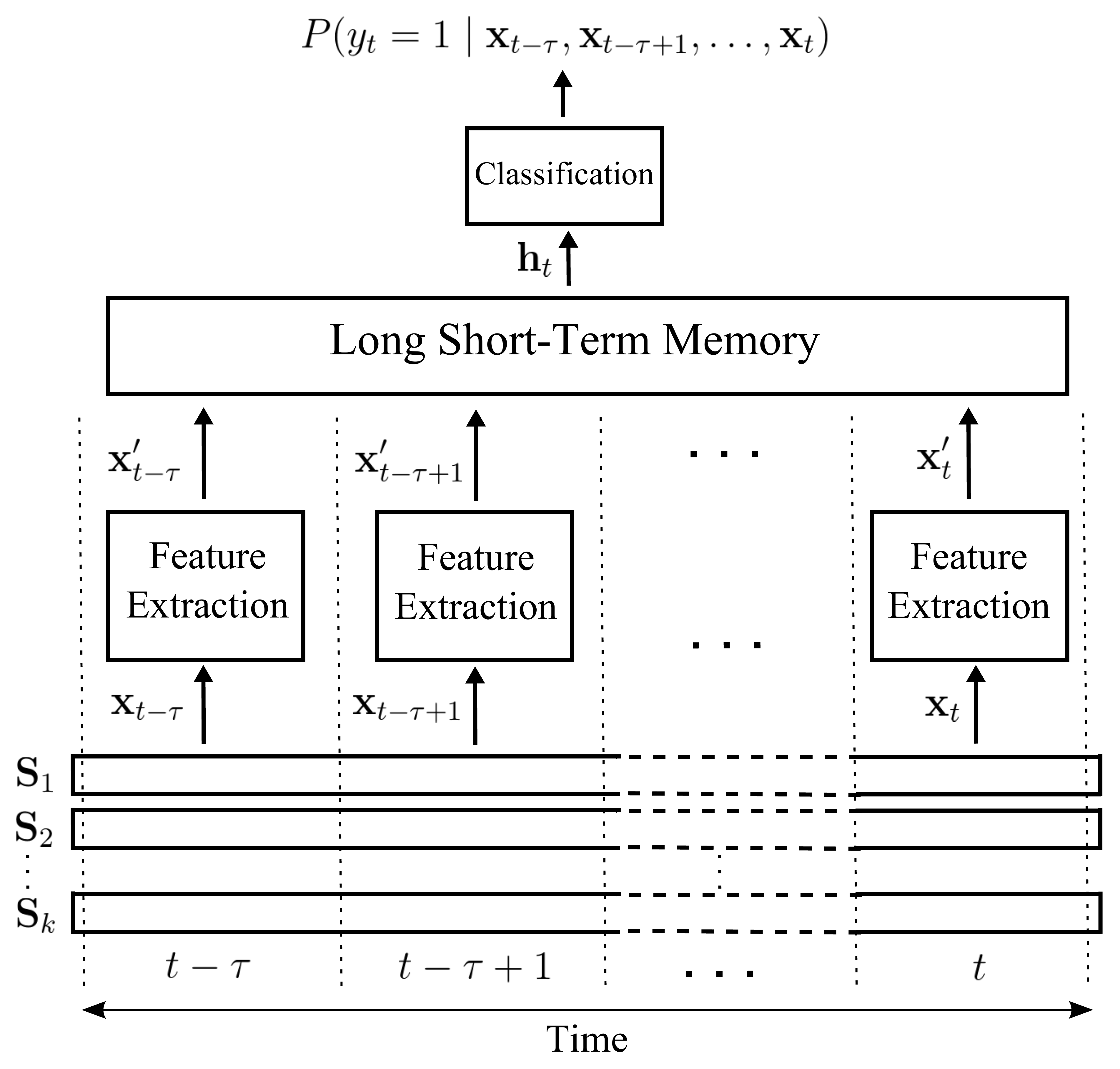}
         \caption{The proposed architecture for SMM detection in the dynamic feature space using long short-term memory. Each feature extraction block contains a trained three-layer CNN architecture (see Figure~\ref{fig:CNN_Architecture}).} 
         \label{fig:architecture}
\end{figure}

In this paper, a fully-connected layer with dropout of $0.2$ is used to transfer the output of the LSTM layer at time $t$, i.e., $\textbf{h}_t$, to the input of the softmax layer $\textbf{z}_t = [z_t^{(0)},z_t^{(1)}]^T \in \mathbb{R}^2$: 
\begin{eqnarray} \label{eq:softmax}
P(y_t=1 \mid \textbf{x}_{t-\tau},\textbf{x}_{t-\tau+1}, \dots, \textbf{x}_{t}) = \frac{e^{z_t^{(1)}}}{e^{z_t^{(0)}}+e^{z_t^{(1)}}} ,
\end{eqnarray}
where $\tau$ represents the number of previous time steps that are used as the input to the LSTM layer. Figure~\ref{fig:architecture} presents a schematic overview of the proposed architecture.

\subsection{Ensemble of the Best Base Learners}
\label{subsec:ensemble_learning}
Due to the random initialization and using stochastic optimization algorithms on random mini-batches in training deep learning models, retraining the same model on the same training set results in heterogeneous approximations of the target classifier. This heterogeneity is the direct result of reaching different local optimums in optimizing a complex non-convex error surface. One possible approach to overcome this problem is \emph{ensemble learning} (EL)~\cite{dietterich2002ensemble}. The main idea behind EL is to combine the knowledge learned by individual classifiers in order to achieve a more superior and stable performance. It is shown that in general an ensemble of classifiers works better than every single classifier due to the statistical, computational, and representational reasons~\cite{kuncheva2004combining}. Considering the success of deep learning ensembles in pattern recognition and signal processing applications~\cite{deng2014ensemble,jin2016ensemble,guan2017ensembles,krawczyk2017ensemble}, in this study we are interested in applying classifier selection voting approach~\cite{gomes2017survey} to combine an ensemble of the best base learners.

Let $(\mathbf{X}_{tr},\mathbf{y}_{tr})$ and $(\mathbf{X}_{ts},\mathbf{y}_{ts})$ to be the corresponding sample/target pairs in the training and test sets, respectively. Then assume $\mathcal{C} = \{c_1, c_2, \dots c_l\}$ be a set of $l$ base learners trained on the training set. Our goal is to first find a set of $b$ best classifiers $\mathcal{C}^* \subseteq \mathcal{C}$ based on a performance measure $\alpha$ on the training set, and then to combine their prediction on the test set using majority voting in the prediction phase. Algorithm~\ref{alg:ensemble} summarizes this approach.
\begin{algorithm}[t]
\caption{The training and test procedures in the majority voting on a set of $b$ best models.} 
\label{alg:ensemble}
\scriptsize
\begin{algorithmic}[1]
\Procedure{training($\mathcal{C}$,$\mathbf{X}_{tr}$,$\mathbf{y}_{tr}$)}{}
	\ForAll{$c_i \in \mathcal{C}$}		
		\State Predict $\hat{\mathbf{y}}$ using $c_i$ on $\mathbf{X}_{tr}$.
		\State Evaluate $\hat{\mathbf{y}}$ and store the performance in $\alpha_i$.
	\EndFor{\textbf{end}}
	\For{$i \gets 1, l$}
		\State Store the best $i$ classifiers in $\mathcal{C}^*_i$.
		\State Predict $\hat{\mathbf{y}}_1, \dots, \hat{\mathbf{y}}_i$ using classifiers in $\mathcal{C}^*_i$ on $\mathbf{X}_{tr}$.
		\State Compute majority voting $\tilde{\mathbf{y}}$ on predictions in $\hat{\mathbf{y}}_1, \dots, \hat{\mathbf{y}}_i$.
		\State Evaluate $\tilde{\mathbf{y}}$ and store the performance in $\alpha_i$.
	\EndFor{\textbf{end}}
	\State Find the best value for $b$ by $b =  \argmax_{i} (\alpha_i)$.
	\State return $\mathcal{C}^*_b$.
\EndProcedure
\\
\Procedure{test($\mathcal{C}^*_b$,$\mathbf{X}_{ts}$, $\mathbf{y}_{ts}$)}{}
	\State Predict $\hat{\mathbf{y}}_1, \dots, \hat{\mathbf{y}}_b$ using classifiers in $\mathcal{C}^*_b$ on $\mathbf{X}_{ts}$.
	\State Compute majority voting $\tilde{\mathbf{y}}$ on predictions in $\hat{\mathbf{y}}_1, \dots, \hat{\mathbf{y}}_b$.
	\State Evaluate $\tilde{\mathbf{y}}$ and store the performance in $\alpha$.
	\State return $\alpha$.
\EndProcedure
\end{algorithmic}
\end{algorithm}

\subsection{Experimental Materials}
\label{subsec:materials}
We assess the performance of the proposed methods on both simulated and real data. In the following, we describe the datasets and the procedures that are used for data preparation.

\subsubsection{Simulated Data}
\label{subsubsec:simulated_data}
\begin{figure*}[t]
    	\centering
        \includegraphics[width=0.95\textwidth]{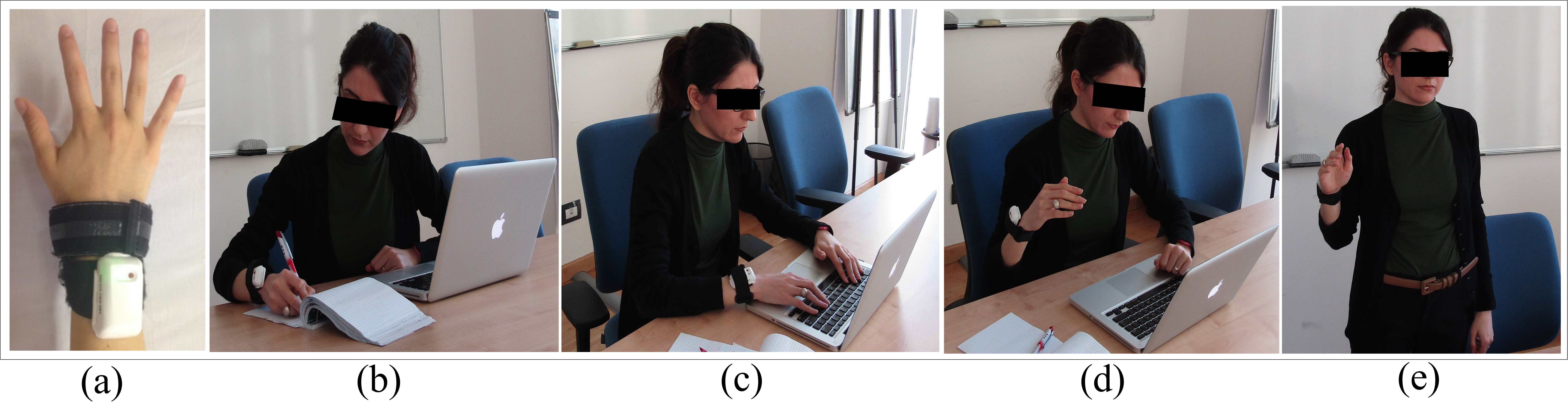}
        \caption{(a) The configuration of the EXLs3 sensor on the right hand. (b),(c) The simulated data are collected during daily work activities (e.g., writing, typing, etc.). (d),(e) The subjects are asked to intermittently perform hand flapping upon receiving a start/stop cue from the instructor.} 
        \label{fig:data_collection}
\end{figure*}

In a simulation setting, 5 healthy subjects (3 females and 2 males) are asked to emulate stereotypical movements in a controlled environment. Each participant wore an EXLs3 sensor\footnote{For the technical description see: \url{http://www.exelmicroel.com/elettronica_medicale-tecnologia-indossabile-exl-s3_module.html}.}, a miniaturized electronic device with the function of real-time IMU, fixed on the right wrist using a wristband (see Figure~\ref{fig:data_collection}(a)). EXLs3 sensor records three-axis accelerometer, gyroscope, and magnetometer data (it has 9 data channels in total). The sensor was set to transmit three-axis $\pm 16g$ acceleration and $\pm 2000dps$ angular velocity at the $100Hz$ sampling rate. The participants were instructed to perform their normal working activities such as sitting, writing, and typing; while intermittently performing hand flapping upon receiving a start/stop cue from the instructor (see Figure~\ref{fig:data_collection}(b)-(e)). The total period of SMMs is organized somehow to keep the distribution of two classes comparable with real datasets where $27\%$ of samples are in the SMM class (see Table~\ref{tab:data} and Section~\ref{subsubsec:real_data}). The total duration of each experiment was $30$ minutes organized in three $10$ minutes sessions. Real-time coding is undertaken during sessions to annotate the starting and ending time of movements. The captured data were band-pass filtered with a cut-off frequency of $0.1Hz$ to remove the DC components. Then the signal was segmented to $1$ second long (i.e., $100$ time-points) using a sliding window. The sliding window was moved along the time dimension with $10$ time-steps resulting in $0.9$ overlaps between consecutive windows~\footnote{The collected simulated data is made publicly available at~\url{https://gitlab.fbk.eu/MPBA/smm-detection}.}.

\subsubsection{Real Data}
\label{subsubsec:real_data}
We use the data presented in~\cite{goodwin2014moving} wherein the accelerometer data were collected for $6$ subjects with autism in a longitudinal study\footnote{The dataset and full description of data are publicly available at \url{https://bitbucket.org/mhealthresearchgroup/stereotypypublicdataset-sourcecodes/downloads}.}. The data were collected in the laboratory and classroom environments while the subjects wore three 3-axis wireless accelerometers and engaged in body rocking, hand flapping, or simultaneous body rocking and hand flapping. The accelerometer sensors were attached to the left and right wrists, and on the torso. Offline annotation based on a recorded video is used to annotate the data by an expert. Two separate collections are available: the first collection, here we call it \emph{Real Data1}, was recorded by MITes sensors at $60Hz$ sampling frequency~\cite{albinali2009recognizing}. The second collection \emph{Real Data2}, was recorded three years after the first recording on the same subjects using Wockets sensors with the sampling frequency of $90Hz$. The sampling rate of two recordings is equalized by re-sampling the signal in \emph{Real Data1} to $90Hz$ using linear interpolation. The cut-off high pass filter at $0.1Hz$ is applied in order to remove the DC components of the signal. Similar to~\cite{goodwin2014moving}, the signal is segmented to 1-second long overlapped intervals using a sliding window. The amount of overlap is set to 10 time-points resulting in $0.87$ overlap between consecutive windows. Table~\ref{tab:data} summarizes the number of samples in no-SMM and SMM classes for each subject. The difference in the number of samples in SMM and no-SMM classes shows unbalanced nature of the real data, where in Real Data1 and Real Data2 datasets $31\%$ and $23\%$ of samples are in the SMM class, respectively.
\begin{table}[t]
\centering
\caption{Number of samples in SMM and no-SMM classes in three datasets.}
\label{tab:data}
\resizebox{0.45\textwidth}{!}{
\begin{tabular}{cccccc}
\hline
\textbf{Data} & \textbf{Subjects} & \textbf{No-SMM} & \textbf{SMM} & \textbf{All} & \textbf{SMM/All} \\ \hline
\multicolumn{1}{c|}{\multirow{6}{*}{\textbf{Simulated Data}}} & \textbf{Sub1} & 13875 & 4075 & 17950 & 0.23 \\
\multicolumn{1}{c|}{} & \textbf{Sub2} & 11686 & 6224 & 17910 & 0.35 \\
\multicolumn{1}{c|}{} & \textbf{Sub3} & 13694 & 4246 & 17940 & 0.24 \\
\multicolumn{1}{c|}{} & \textbf{Sub4} & 12428 & 5532 & 17960 & 0.31 \\
\multicolumn{1}{c|}{} & \textbf{Sub5} & 13583 & 4367 & 17950 & 0.24 \\
\multicolumn{1}{c|}{} & \textbf{Total} & 65266 & 24444 & 89710 & 0.27 \\ \hline
\multicolumn{1}{c|}{\multirow{7}{*}{\textbf{Real Data1}}} & \textbf{Sub1} & 21292 & 5663 & 26955 & 0.21 \\
\multicolumn{1}{c|}{} & \textbf{Sub2} & 12763 & 4372 & 17135 & 0.26 \\
\multicolumn{1}{c|}{} & \textbf{Sub3} & 31780 & 2855 & 34635 & 0.08 \\
\multicolumn{1}{c|}{} & \textbf{Sub4} & 10571 & 10243 & 20814 & 0.49 \\
\multicolumn{1}{c|}{} & \textbf{Sub5} & 17782 & 6173 & 23955 & 0.26 \\
\multicolumn{1}{c|}{} & \textbf{Sub6} & 12207 & 17725 & 29932 & 0.59 \\
\multicolumn{1}{c|}{} & \textbf{Total} & 106395 & 47031 & 153426 & 0.31 \\ \hline
\multicolumn{1}{c|}{\multirow{7}{*}{\textbf{Real Data2}}} & \textbf{Sub1} & 18729 & 11656 & 30385 & 0.38 \\
\multicolumn{1}{c|}{} & \textbf{Sub2} & 22611 & 4804 & 27415 & 0.18 \\
\multicolumn{1}{c|}{} & \textbf{Sub3} & 40557 & 268 & 40825 & 0.01 \\
\multicolumn{1}{c|}{} & \textbf{Sub4} & 38796 & 8176 & 46972 & 0.17 \\
\multicolumn{1}{c|}{} & \textbf{Sub5} & 22896 & 6728 & 29624 & 0.23 \\
\multicolumn{1}{c|}{} & \textbf{Sub6} & 2375 & 11178 & 13553 & 0.82 \\
\multicolumn{1}{c|}{} & \textbf{Total} & 145964 & 42810 & 188774 & 0.23 \\ \hline
\end{tabular}}
\end{table}

\subsection{Experimental Setups and Evaluation}
\label{subsec:setups}
To investigate the effect of static and dynamic feature learning and parameter transfer learning on the performance of SMM detection, we conducted four experiments. Keras library~\cite{chollet2015keras} is used in our implementations\footnote{See~\url{https://gitlab.fbk.eu/MPBA/smm-detection} to access the implemented scripts and codes.}. 

\subsubsection{Experiment 1: Static Feature Learning} 
The main aim of this experiment is to compare the effectiveness of feature learning using a deep neural network versus raw feature space and handcrafted features in an across-subject SMM detection setting. To evaluate the effect of both feature extraction and feature learning on the SMM classification performance, first, without any feature extraction the signals in raw feature space are used as the input to a support vector machine (SVM) classifier. In this case, all data channels of each sample $\textbf{X}_t$ are collapsed into a feature vector (with length of $900 = 9 \times 100$ in simulated data and $810 = 9 \times 90$ in real data case). Second, to evaluate the detection performance using handcrafted features we extracted all features mentioned in~\cite{goodwin2014moving} including time, frequency, and Stockwell transform features, then, we replicated the across-subject SMM detection experiment in~\cite{goodwin2014moving}. In this setting we used exactly the same implementation provided by the authors~\footnote{The code is available at: \url{https://bitbucket.org/mhealthresearchgroup/stereotypypublicdataset-sourcecodes/downloads}.} in the feature extraction and classification steps. Third, a CNN architecture (see Section~\ref{subsec:CNN}) is used to learn a middle representation of the multi-sensor signal. In this experiment, all effective parameters of CNN (weights and biases) are initialized by drawing small random numbers from the normal distribution. The stochastic gradient descent with momentum (the momentum is fixed to $0.9$) is used for training the network. All these steps are performed only on the training data to ensure unbiased error estimation. Due to the random initialization of weights and employing stochastic gradient descent algorithm for optimization, results can be different from one training run to another. Therefore, we repeated the whole procedure of learning and classification $10$ times and the mean and standard variation over runs are reported. It is important to emphasize that, similar to~\cite{goodwin2014moving}, in all three parts of this experiment the number of samples in minority class is used to randomly draw a balanced training set. 

\subsubsection{Experiment 2: Parameter Transfer Learning} 
\label{subsubsec:exp2}
As discussed before, deep neural networks provide the capability of parameter transfer learning via network pre-initialization. We applied this experiment only on two real datasets in order to investigate the possibility of transferring learned knowledge from one dataset to another in a longitudinal data collection setting. This experiment is similar to Experiment 1, except for the network initialization step. Instead of random initialization, here we firstly train the CNN on one balanced real dataset, e.g., Real Data1, and then we use the learned parameters for pre-initializing the parameters of CNN before training on another balanced real dataset, e.g., Real Data2. Similar to previous experiment, we repeated the whole experiment $10$ times to evaluate the standard deviation of the classification performance.
\begin{figure*}[th!]
         \centering
         \includegraphics[width=0.95\textwidth]{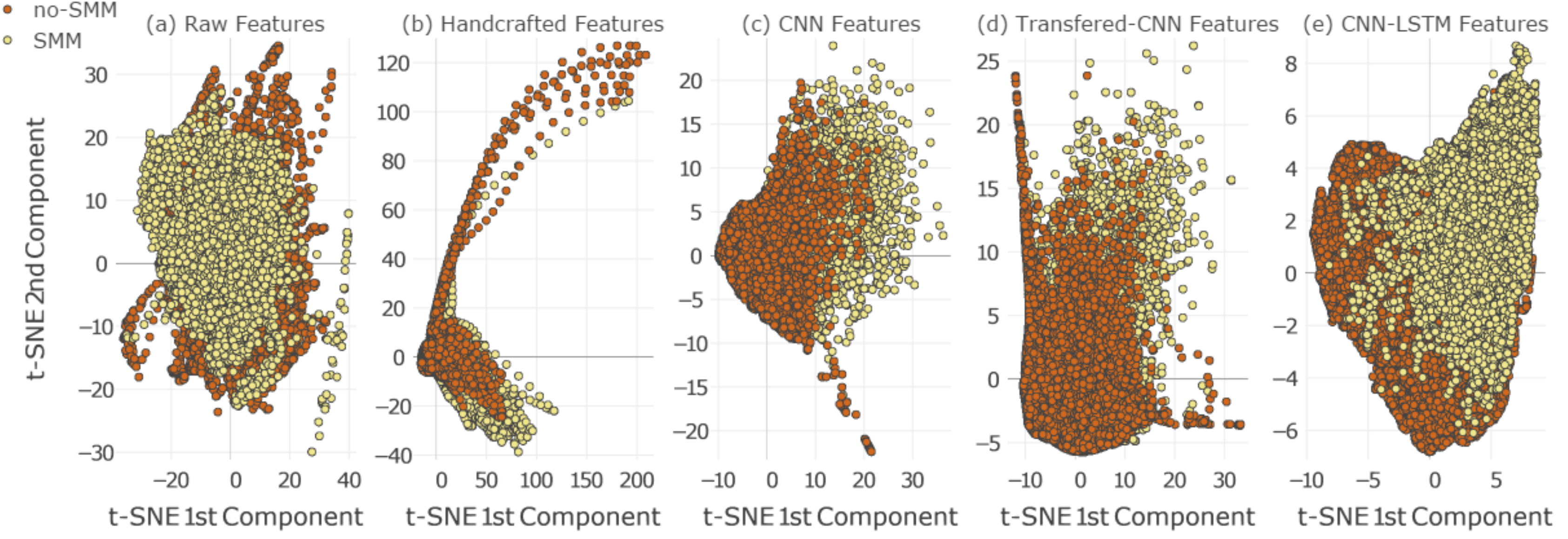}
         \caption{The distribution of SMM and no-SMM samples in the 2-dimensional t-SNE space for (a) raw feature space, (b) handcrafted features, (c) static feature space learned by CNN, (d) static feature space learned by pre-initialized CNN, and (e) dynamic feature space learned by LSTM. Feature learning increases the separability of samples in two classes compared to raw and handcrafted features.} 
         \label{fig:tsne_scatter}
\end{figure*}

\subsubsection{Experiment 3: Training on the Unbalanced Training Set}
As explained, in Experiment 1 and 2 we balanced the training set based on the number of samples in minority class. Even though balancing the training set improves the quality of the trained model but in fact it suffers from some deficits: 1) by balancing the training set we impose a wrong prior assumption on the original distribution of data. As shown in Table~\ref{tab:data} in real datasets around 0.3 of samples belong to SMM class, when by balancing the dataset we assume it is 0.5; 2) by balancing the training set we cannot employ the full richness of the data as we need to remove significant amount of samples from the training set; 3) in some practical scenarios, such as real-time adaptation or classification on the sequence of streamed data, balancing the training set is impractical. Considering these limitations, in this experiment in order to evaluate the effect of balancing on the performance of CNN model we evaluate the performance of the proposed CNN architecture in predicting SMMs when unbalanced training sets are used in the training phase.

\subsubsection{Experiment 4: Dynamic Feature Learning}
In this experiment, we are interested in answering three main questions: 1) what are the advantages of learning temporal representation of IMU signals for reliable SMM detection? 2) how long is the most informative time interval in IMU signals for detecting abnormal movements? 3) what is the optimal configuration for the LSTM unit? To answer these questions, we applied the proposed LSTM architecture in Section~\ref{subsec:dynamic_SMM_detection} on the three benchmark datasets with different values for $\tau$ and $q$, i.e., time steps and neuron number, respectively. We set $\tau=\{1,3,5,10,15,25,50\}$ and $q = \{5,10,20,30,40,50\}$. The LSTM unit is trained on the extracted features by the CNN using the RMSProp~\cite{tieleman2012lecture} optimizer. The learned dynamic features via LSTM ($\textbf{h}_t$) are classified to target classes using a softmax classifier. It is worthwhile to emphasize that, in this setting, since the order of samples in the training set matters, balancing the training set is impossible, thus we use the original unbalanced data.

\subsubsection{Experiment 5: Ensemble of LSTMs}
To explore the possible advantage of combining multiple classifiers, we used the procedure explained in Section~\ref{subsec:ensemble_learning} in order to combine a set of $b$ best base learners. In this experiment, we used the LSTM models in the Experiment 4 as base learners for the classification of unbalanced data. For all datasets, we used the same configurations for the LSTM models by fixing $\tau = 25$ and $q = 40$. We set $l = 10$ and used the F1-score for the performance metric $\alpha$ in Algorithm~\ref{alg:ensemble}. The experiment is repeated $10$ times to evaluate the standard deviation over the mean prediction performance.

\subsubsection{Evaluation} 
In all experiments the leave-one-subject-out scheme is used for model evaluation in order to measure the robustness of the trained model against inter-subject variability. Due to the unbalanced class distributions in the test set, we computed the F1-score to evaluate the classification performance:
\begin{eqnarray*} \label{eq:f1_score}
F1 = 2 \times \frac{Precision \times Recall}{Precision + Recall} \quad ,
\end{eqnarray*}
where true positive (TP), false positive (FP), and false negative (FN) rates are used as follows to compute the precision and recall:
\begin{align*} \label{eq:Pre_Rec}
Precision &= \frac{TP}{TP + FP} \\
Recall &= \frac{TP}{TP + FN} .
\end{align*}

\begin{table*}[t]
\centering
\caption{Results for SMM detection using raw, handcrafted, static, dynamic feature spaces, and ensemble learning in three benchmarked datasets. The results show that feature learning generally outperforms raw and handcrafted feature spaces. In addition, the parameter transfer learning has a positive effect on the performance of the CNN classifier. Furthermore, training the CNN classifier on unbalanced training sets causes the performance drop in feature learning and transfer learning scenarios. Using the LSTM network to extract dynamic features from the signal alleviates this problem to some degrees. Ensemble of LSTMs shows more stable performance compared to single LSTM classifiers.}
\label{tab:results}
\resizebox{1\textwidth}{!}{\begin{tabular}{cc|cccc|ccccc}
\multirow{2}{*}{\textbf{Data}} & \textbf{} & \multicolumn{4}{c|}{\textbf{Balanced Training Sets}} & \multicolumn{5}{c}{\textbf{Unbalanced Training Sets}} \\ \cline{3-11} 
 & \textbf{Sub} & \textbf{\begin{tabular}[c]{@{}c@{}}Raw \\ Features\end{tabular}} & \textbf{\begin{tabular}[c]{@{}c@{}}Handcrafted\\ Features\end{tabular}} & \textbf{\begin{tabular}[c]{@{}c@{}}Feature\\ Learning\end{tabular}} & \textbf{\begin{tabular}[c]{@{}c@{}}Transfer \\ Learning\end{tabular}} & \textbf{\begin{tabular}[c]{@{}c@{}}Feature\\ Learning\\ (1 sec)\end{tabular}} & \textbf{\begin{tabular}[c]{@{}c@{}}Transfer\\ Learning\end{tabular}} & \textbf{\begin{tabular}[c]{@{}c@{}}Feature\\ Learning\\  (2.5 sec)\end{tabular}} & \textbf{\begin{tabular}[c]{@{}c@{}}Dynamic \\ Feature\\ Learning\end{tabular}} & \textbf{\begin{tabular}[c]{@{}c@{}}Ensemble \\ Learning\end{tabular}} \\ \hline
\multirow{6}{*}{\textbf{\begin{tabular}[c]{@{}c@{}} \rot{Simulated} \end{tabular}}} & \textbf{1} & 0.29 & 0.71 & $0.78\pm0.05$ & - & $0.73\pm0.13$ & - & $0.95\pm0.01$ & $0.95\pm0.01$ & $0.95\pm0.00$ \\
 & \textbf{2} & 0.84 & 0.86 & $0.86\pm0.03$ & - & $0.78\pm0.09$ & - & $0.86\pm0.13$ & $0.95\pm0.01$ & $0.96\pm0.01$ \\
 & \textbf{3} & 0.55 & 0.76 & $0.80\pm0.01$ & - & $0.75\pm0.13$ & - & $0.95\pm0.03$ & $0.97\pm0.01$ & $0.97\pm0.00$ \\
 & \textbf{4} & 0.76 & 0.48 & $0.85\pm0.03$ & - & $0.73\pm0.12$ & - & $0.97\pm0.01$ & $0.96\pm0.02$ & $0.97\pm0.01$ \\
 & \textbf{5} & 0.38 & 0.77 & $0.79\pm0.01$ & - & $0.80\pm0.04$ & - & $0.91\pm0.01$ & $0.90\pm0.02$ & $0.91\pm0.00$ \\
 & \textbf{Mean} & 0.56 & 0.72 & $0.82\pm0.03$ & - & $0.76\pm0.11$ & - & $0.93\pm0.06$ & $0.95\pm0.01$ & $0.95\pm0.01$ \\ \hline
\multirow{7}{*}{\textbf{\begin{tabular}[c]{@{}c@{}} \rot{Real Data1} \end{tabular}}} & \textbf{1} & 0.44 & 0.74 & $0.74\pm0.02$ & $0.71\pm0.02$ & $0.70\pm0.02$ & $0.71\pm0.03$ & $0.73\pm0.04$ & $0.77\pm0.03$ & $0.80\pm0.00$ \\
 & \textbf{2} & 0.32 & 0.37 & $0.75\pm0.02$ & $0.73\pm0.01$ & $0.63\pm0.03$ & $0.63\pm0.04$ & $0.68\pm0.04$ & $0.71\pm0.03$ & $0.74\pm0.00$ \\
 & \textbf{3} & 0.22 & 0.50 & $0.68\pm0.04$ & $0.70\pm0.03$ & $0.57\pm0.08$ & $0.59\pm0.06$ & $0.56\pm0.13$ & $0.68\pm0.05$ & $0.72\pm0.01$ \\
 & \textbf{4} & 0.44 & 0.73 & $0.92\pm0.01$ & $0.92\pm0.00$ & $0.88\pm0.01$ & $0.88\pm0.01$ & $0.93\pm0.00$ & $0.91\pm0.01$ & $0.93\pm0.00$ \\
 & \textbf{5} & 0.56 & 0.44 & $0.51\pm0.04$ & $0.68\pm0.05$ & $0.51\pm0.08$ & $0.58\pm0.07$ & $0.51\pm0.04$ & $0.52\pm0.04$ & $0.51\pm0.01$ \\
 & \textbf{6} & 0.56 & 0.46 & $0.90\pm0.01$ & $0.94\pm0.01$ & $0.79\pm0.07$ & $0.81\pm0.09$ & $0.86\pm0.12$ & $0.90\pm0.02$ & $0.91\pm0.00$ \\
 & \textbf{Mean} & 0.42 & 0.54 & $0.74\pm0.03$ & $0.78\pm0.03$ & $0.68\pm0.06$ & $0.70\pm0.06$ & $0.71\pm0.08$ & $0.75\pm0.03$ & $0.77\pm0.01$ \\ \hline
\multirow{7}{*}{\textbf{\begin{tabular}[c]{@{}c@{}} \rot{Real Data2} \end{tabular}}} & \textbf{1} & 0.47 & 0.43 & $0.61\pm0.11$ & $0.68\pm0.05$ & $0.33\pm0.14$ & $0.36\pm0.08$ & $0.47\pm0.15$ & $0.53\pm0.09$ & $0.59\pm0.03$ \\
 & \textbf{2} & 0.23 & 0.26 & $0.20\pm0.04$ & $0.22\pm0.04$ & $0.11\pm0.03$ & $0.16\pm0.04$ & $0.13\pm0.05$ & $0.26\pm0.06$ & \multicolumn{1}{l}{$0.29\pm0.02$} \\
 & \textbf{3} & 0.01 & 0.03 & $0.02\pm0.01$ & $0.02\pm0.01$ & $0.02\pm0.01$ & $0.02\pm0.01$ & $0.02\pm0.01$ & $0.02\pm0.02$ & \multicolumn{1}{l}{$0.02\pm0.02$} \\
 & \textbf{4} & 0.32 & 0.86 & $0.72\pm0.03$ & $0.77\pm0.02$ & $0.71\pm0.14$ & $0.83\pm0.03$ & $0.90\pm0.02$ & $0.76\pm0.09$ & \multicolumn{1}{l}{$0.87\pm0.02$} \\
 & \textbf{5} & 0.38 & 0.73 & $0.21\pm0.09$ & $0.75\pm0.09$ & $0.14\pm0.09$ & $0.09\pm0.02$ & $0.23\pm0.17$ & $0.35\pm0.15$ & \multicolumn{1}{l}{$0.43\pm0.08$} \\
 & \textbf{6} & 0.50 & 0.07 & $0.36\pm0.13$ & $0.91\pm0.05$ & $0.62\pm0.08$ & $0.70\pm0.16$ & $0.68\pm0.21$ & $0.96\pm0.01$ & \multicolumn{1}{l}{$0.98\pm0.00$} \\
 & \textbf{Mean} & 0.32 & 0.40 & $0.35\pm0.08$ & $0.56\pm0.05$ & $0.32\pm0.1$ & $0.36\pm0.08$ & $0.40\pm0.13$ & $0.48\pm0.08$ & \multicolumn{1}{l}{$0.53\pm0.04$} \\ \hline
\end{tabular}}
\end{table*}

\section{Results}
\label{sec:results}
\subsection{Feature Learning Outperforms Handcrafted Features}
\label{subsec:feature_learning_results}
The classification performances summarized in Table~\ref{tab:results} compare the quality of feature learning via CNN with raw and handcrafted feature spaces on three datasets. In all three datasets, the classification performance of SMM detection on the handcrafted and learned features is higher than the classification performance on the raw feature space. This observation demonstrates the importance of feature extraction/learning for detecting SMMs. Furthermore, the comparison between the results achieved by handcrafted and learned features illustrates the efficacy of feature learning over the manual feature extraction in SMM prediction. The learned feature space reaches on average $0.10$ and $0.20$ higher F1-score than the handcrafted features in case of simulated data and real data1, respectively, while in case of real data2 its performance declines by $0.05$. These results support the overall efficacy of feature learning versus handcrafted features in extracting robust features for across-subject SMM detection. These conclusions are even further confirmed in Figures~\ref{fig:tsne_scatter}(a)-(c), where the t-distributed Stochastic Neighbor Embedding (t-SNE)~\cite{maaten2008visualizing} technique is employed to visualize the different feature spaces in a 2-dimensional space. We used the average of Fisher's separability score~\cite{duda2012pattern} across two t-SNE dimensions to quantify the separability of samples in two classes for different feature spaces. Figure~\ref{fig:tsne_scatter}(a) shows 2D t-SNE distribution of SMM and no-SMM samples in the raw feature space, where there is a high overlap between the samples of two classes. This high overlap is also well-reflected in the low Fisher's separability score in raw feature space ($0.02$). Figure~\ref{fig:tsne_scatter}(b) depicts the distribution of samples of two classes in handcrafted feature space. The samples in two classes are barely separable and the Fisher's separability score is $0.03$. Figure~\ref{fig:tsne_scatter}(c) displays the 2D t-SNE space for the learned features via the CNN architecture. In this case the separability score is improved significantly to $0.10$.

\subsection{Parameter Transfer Learning is Beneficial in Longitudinal Studies}
\label{subsec:longitudinal_transfer_learning}
As mentioned in Section~\ref{subsubsec:exp2}, the aim of our second experiment was to investigate the possibility of transferring learned knowledge from one dataset to another using parameter transfer learning. Our results in Table~\ref{tab:results} shows that transferring knowledge from one dataset to another in a longitudinal study, by pre-initializing the parameters of CNN model improves the average classification performance of the SMM detectors by $0.04$ and $0.21$ in Real Data1 and Real Data2 datasets, respectively. 

\subsection{Training on Unbalanced Data Decreases the Performance}
\label{subsec:unbalanced_Data}
The results in Table~\ref{tab:results} illustrate the negative effect of using unbalanced training set in training CNN architecture in randomly initialized (feature learning) and pre-initialized (transfer learning) scenarios. The performance of SMM detection in feature learning scenario drops by $0.06$ and $0.03$ in Real Data1 and Real Data2 datasets, respectively. This performance drop is even more pronounced in the transfer learning scenario where we observe $0.08$ and $0.20$ performance drop in the corresponding datasets. 
\begin{figure*}[th!]
         \centering
         \includegraphics[width=0.95\textwidth]{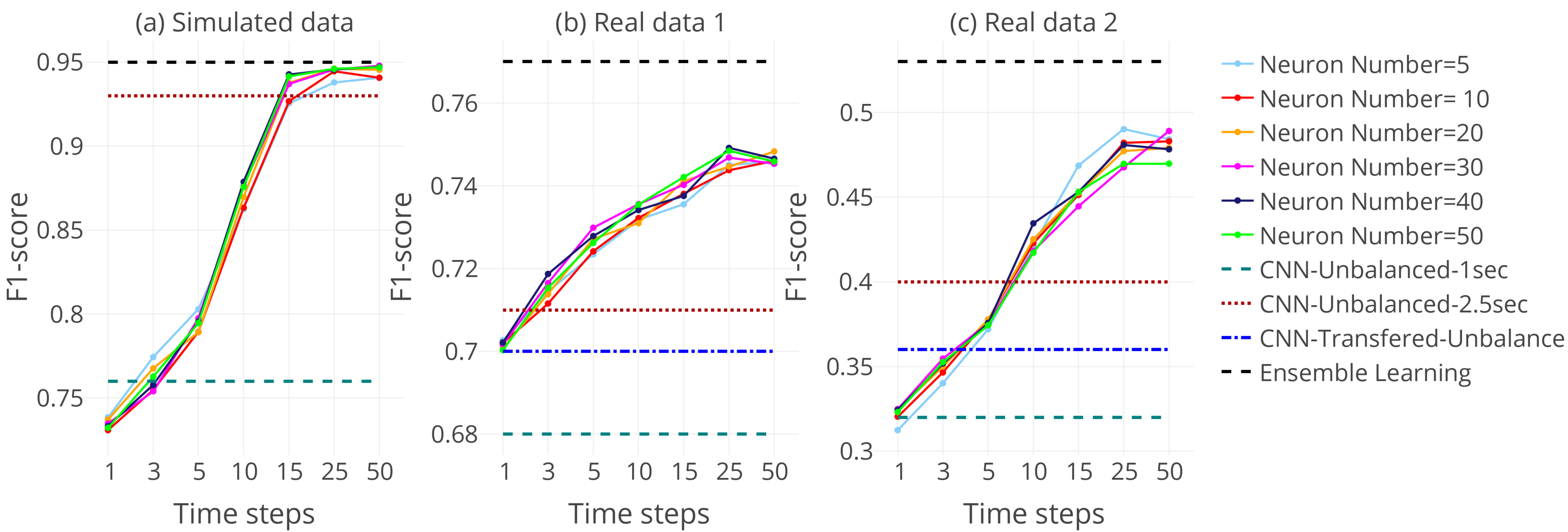}
         \caption{Comparison between the classification performances of CNN and LSTM for different time-steps ($\tau$) and number of neurons $q$, on three datasets and when an unbalanced training set is used for training the networks. The results show the superiority of the dynamic feature space over the static feature space. While the number of neurons in the LSTM unit has little effect on the performance, using around $2.5$ seconds long interval is the best choice for extracting effective dynamic features from the IMU signals.} 
         \label{fig:time_steps}
\end{figure*}

\subsection{Dynamic Feature Learning Outperforms Static Feature Learning}
\label{subsec:dynamic_feature_learning}
Figure~\ref{fig:time_steps} compares the averaged SMM classification performance over subjects in the static feature space via the CNN (the green dashed line for plain feature learning and the blue dotted-dashed line for transfer learning) with the dynamic feature space via the LSTM, the latter with different values for $\tau$ (x-axis) and $q$ (line colors). Here in all settings, an unbalanced training set is used in the training phase. The results on three datasets illustrate that learning the temporal representation of signals with an LSTM unit, consistently across datasets, improves the classification performance compared to the static feature learning via the CNN. The classification performance improves by increasing $\tau$, and it reaches its highest performance around $\tau=25$. Considering the consistency of the best $\tau$ value for different subjects and different datasets, it can be concluded that using around $25$ time-steps, i.e., around $2.5$ seconds long interval, for extracting dynamic features is the best choice for SMM detection purposes. On the other hand, the results show the negligible effect of the number of LSTM neurons ($q$) on the detection performance, thus, a value around $10$ can be considered a reasonable choice.

To further benchmark the advantage of dynamic feature learning via LSTM, we used the CNN architecture for the SMM detection on the best length for the time intervals, i.e., on $2.5$ seconds time intervals. The results on the three datasets are summarized in Table~\ref{tab:results} and Figure~\ref{fig:time_steps} (the dotted red line). The results confirm the superiority of dynamic feature learning compared to static feature learning despite using longer time intervals for learning static features.

The effect of learning the temporal representation on the separability of SMM and no-SMM samples is shown in Figure~\ref{fig:tsne_scatter}(e). The higher Fisher's separability score ($0.17$) in the dynamic feature space compared to static feature spaces can be considered as the basis for the higher classification performance of the proposed architecture, demonstrating the importance of learning dynamic features using an LSTM based architecture. Furthermore, employing the dynamic-feature representation computed by the LSTM, improves the detection rates when unbalanced training sets are used (see Table~\ref{tab:results}). This observation is consistent across three datasets. 

\begin{figure*}[th!]
         \centering
         \includegraphics[width=0.95\textwidth]{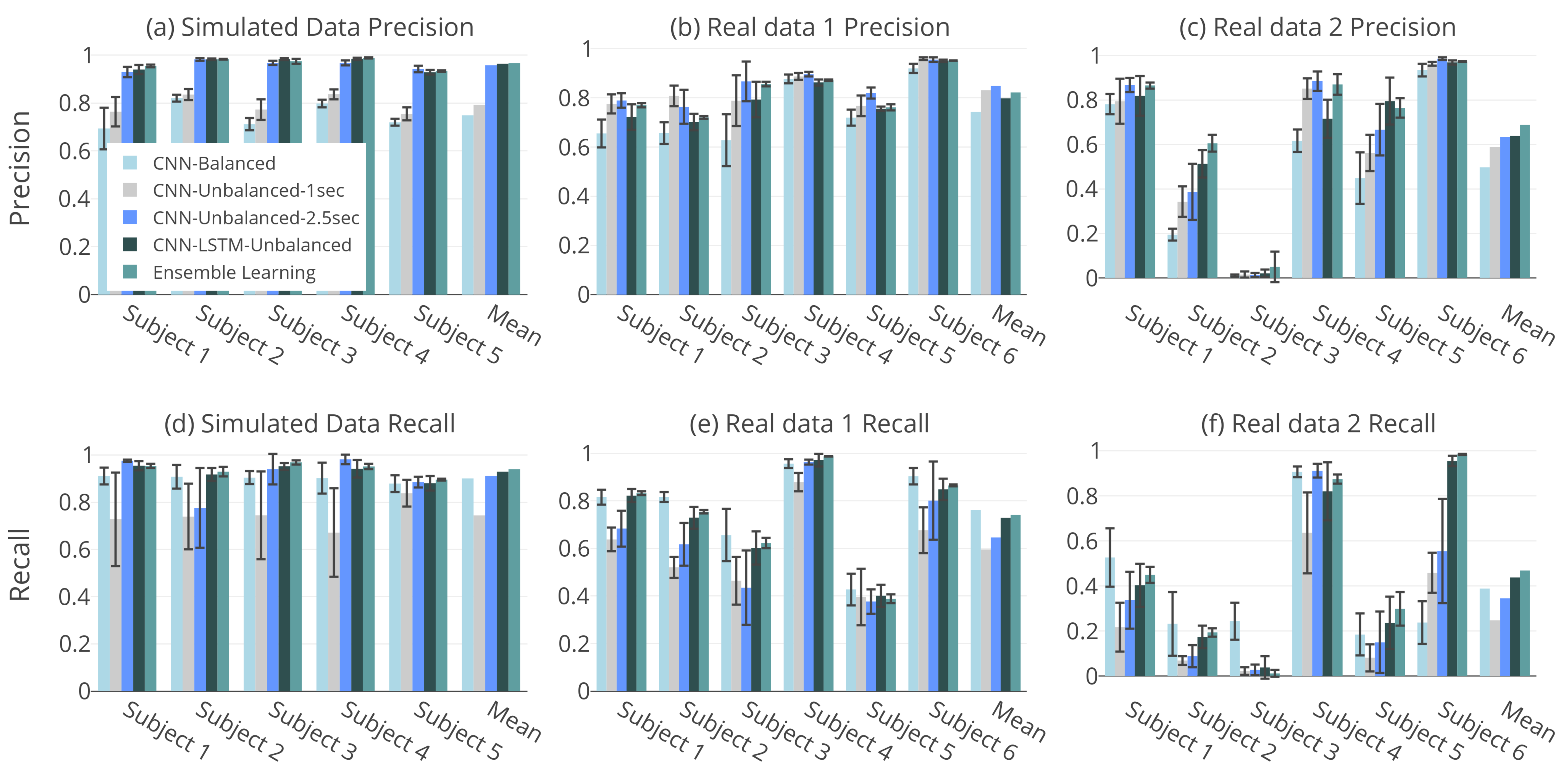}
         \caption{A comparison between precision and recall rates of classifiers trained on balanced/unbalanced training sets for static/dynamic feature representations. Using dynamic feature space provides an SMM detection system with higher sensitivity and specificity.} 
         \label{fig:precision_recall}
\end{figure*}

Figure~\ref{fig:precision_recall} further explores the superiority of the dynamic feature representation when the training set is unbalanced. In the static feature space case, balancing the training set and enforcing the wrong prior class distribution into the classification task, despite higher recall rate, affects negatively the precision of the classifier. In other words, the classifiers have higher false alarm rate, which could be problematic in real-world applications. This deficit is recovered in the case of dynamic feature representation where the classifier presents higher precision rate and comparable recall with respect to static features. In fact, the LSTM-based architecture by enforcing the true prior distribution of data into the training process and, at the same time using all the recorded samples, provides an SMM detection system with higher sensitivity and specificity. 

\subsection{Ensemble of LSTMs Stabilizes the Performance}
\label{subsec:results_ensemble_learning}
The last column of Table~\ref{tab:results} summarizes the results of the ensemble approach. The results show slight boost in the  mean performance compared to single LSTM classifiers, especially on real data2 (see dashed black line in Figure~\ref{fig:time_steps}). Figure~\ref{fig:precision_recall} shows that both precision and recall contribute equally to this improvement in F1-scores. In addition to the higher performance, the main advantage of EL is demonstrated by the low variability of results. This reduction in the variability is well-reflected in the reduced standard deviation around the mean performance in real datasets ($0.02$ and $0.04$ reduction in real data1 and real data2 datasets, respectively). In other words an ensemble of LSTMs provides more reliable SMM detector in comparison to every single LSTM classifier.
\section{Discussion}
\label{sec:discussion}
\subsection{Toward Real-Time Automatic SMM Detection}
\label{subsec:toward_realtime}
In this study, we proposed an original application of deep learning for SMM detection in ASD children using wearable technologies. To this end, we used a three-layer CNN architecture for learning a more discriminative and robust feature space in an across-subject SMM detection scenario. Our experimental results, on the simulated and real data, support the superiority of learning middle representation of IMU signal over traditional feature extraction methods in automatic SMM detection. Further, we showed that the parameter transfer learning via network pre-initialization provides the infrastructure for effective knowledge transfer from one dataset to another in a longitudinal setting. We also presented an application of LSTM in SMM detection context for extracting dynamic features on the sequence of IMU data. In a comparison with the static feature space learned via CNN architecture, we illustrated the advantage of employing the temporal information in improving the separability of SMM and no-SMM samples. We experimentally showed that using around $2.5$ seconds long interval for extracting dynamic features is the best choice for SMM detection purposes. Further, we showed using around $10$ neurons in the LSTM unit is a reasonable choice in order to extract the dynamics of samples over time. As a side-advantage of learning a dynamic feature space, we experimentally demonstrated the higher performance of our method when, in a real world setting, the distribution of samples in SMM and no-SMM classes is highly skewed. We showed, while the skewness of samples negatively affects the performance of the SMM detector in the static feature space, exploiting the temporal patterns of multi-sensor IMU signals recovers its performance. This advantage facilitates training high-performing models by exploiting whole data sequences in real-time SMM detection scenarios. Our effort, for the first time in the SMM detection context, demonstrated the superiority of recurrent structures in extracting discriminative temporal patterns from IMU signals. As the final contribution, considering the important role of ensemble learning for classifying the stream data in non-stationary environments~\cite{gomes2017survey,krawczyk2017ensemble}, we employed ensemble of the best base learners technique to improve the reliability of the SMM detector. In summary, our results show that feature learning, transfer learning, ensemble learning, and learning temporal structures from multi-modal IMU signals improve the performance of SMM detection systems especially in more realistic scenarios.

Developing real-time mobile applications for detecting the abnormal movements such as SMMs can be considered as a final goal in the context of automatic SMM detection using wearable sensors. At the moment there are numerous challenges in real-time human activity recognition using wearable sensors, namely~\cite{lara2013survey,labrador2013human,lockhart2014limitations}: 1) designing effective feature extraction and inference methods; 2) recognition on realistic data; 3) the adaptability of system to new users. Addressing these issues demands a huge investment in research toward finding robust and effective features that can be extracted in a reasonable time from the stream of IMU signal. Our proposal to learn a middle representation of the signal, that is robust to the signal variability of a single subject data over time and also to the across-subject variability, can be considered as an effective solution in this direction. In addition, the parameter transfer learning capability besides the possibility of incremental training of the proposed deep architecture facilitates the online adaptation of an automatic SMM detector in real-time scenarios. This finding overlooks the subject specific~\cite{berchtold2010extensible}, and monolithic~\cite{lara2012centinela,Großekathöfer2017Automated} activity recognition systems opening new frontiers toward adaptable activity recognition systems which are more appropriate for real-time usages. At the end, the high detection performance on unbalanced training sets achieved in the dynamic feature space facilitates the application of our method on the realistic data when the incoming data samples are highly skewed. 

\subsection{Limitation and Future Work}
\label{subsec:limitation_future}
Even though the deep architecture introduced in this study provides a significant step toward a more accurate automatic SMM detection system in real-time scenarios, but it suffers from a considerable limitation: the proposed fully supervised scheme for training the SMM detection model is problematic for its online adaptation. This problem comes from the fact that in real applications the system has no access to the labels of incoming samples during usage by a new user. Therefore, the adaptation to new unseen data should be performed only based on the input unlabeled data. This limitation motivates future researches on the online adaptation of the system in an unsupervised manner. One possible solution in this direction is transductive transfer learning~\cite{pan2010survey} where the basic assumption is that no labeled data in the target domain are available. Therefore adopting a transductive transfer learning strategy in the adaptation phase can be considered as a possible future direction to extend this work.
\section{Conclusions}
\label{sec:conclusions}
In this study we addressed the problem of automatic SMM detection in the framework of deep learning. We used a 3-layer CNN architecture to learn a robust feature space from multi-sensor/modal IMU signals. We illustrated how the proposed architecture can be employed for parameter transfer learning in order to enhance the adaptability of SMM detection system to new data. Further, we showed incorporating the temporal dynamics of the signal in the feature learning process, by combining the CNN architecture with an LSTM unit, improves the SMM detection rate in real-world scenarios especially in case of unbalanced data. We further illustrated the advantage of ensemble learning to provide more stable and reliable SMM detectors. Our results demonstrate high potentials of deep learning paradigm to address the crucial challenges toward real-time SMM detection systems using wearable technologies. 

\section*{Acknowledgments}
\label{sec:acknowledgments}
The authors would like to thank the members of MPBA lab for their kind collaboration in collecting the simulated data.


\bibliographystyle{model1-num-names}
\footnotesize
\bibliography{references}

\end{document}